\documentclass[runningheads]{llncs}
\usepackage{graphicx}

\begin{document}
\title{Leaf: Multiple-Choice Question Generation}
\author{Kristiyan Vachev\inst{1}
\and Momchil Hardalov\inst{1}
\and Georgi Karadzhov\inst{2}
\and Georgi Georgiev\inst{3}
\and Ivan Koychev\inst{1}
\and Preslav Nakov\inst{4}}
\authorrunning{K. Vachev et al.}
\institute{
{FMI, Sofia University ``St. Kliment Ohridski'', Bulgaria}
\and
{University of Cambridge, UK}
\and
{Releva.ai, Bulgaria}
\and
{Qatar Computing Research Institute, HBKU, Qatar}
}
\maketitle              %
\begin{abstract} 
Testing with quiz questions has proven to be an effective way to assess and improve the educational process. However, manually creating quizzes is tedious and time-consuming. 
To address this challenge, we present Leaf, a system for generating multiple-choice questions from factual text.
In addition to being very well suited for the classroom, Leaf could also be used in an industrial setting, e.g.,~to facilitate onboarding and knowledge sharing, or as a component of chatbots, question answering systems, or Massive Open Online Courses (MOOCs). The code and the demo are available on GitHub.\footnote{\url{https://github.com/KristiyanVachev/Leaf-Question-Generation}}

\end{abstract}

\section{Introduction}

Massive Open Online Courses (MOOCs) have revolutionized education by offering a wide range of educational and professional training. However, an important issue in such a MOOC setup is to ensure an efficient student examination setup. Testing with quiz questions has proven to be an effective tool, which can help both learning and student retention~\cite{ROEDIGERIII20111}. Yet, preparing such questions is a tedious and time-consuming task, which can take up to 50\% of an instructor's time~\cite{susanti2017evaluation}, especially when a large number of questions are needed in order to prevent students from memorizing and/or leaking the answers.

To address this issue, we present an automated multiple-choice question generation system with focus on educational text. Taking the course text as an input, the system creates question--answer pairs together with additional incorrect options (distractors).
It is very well suited for a classroom setting, and the generated questions could also be used for self-assessment and for knowledge gap detection, thus allowing instructors to adapt their course material accordingly. It can also be applied in industry, e.g.,~to produce questions to enhance the process of onboarding, to enrich the contents of massive open online courses (MOOCs), or to generate data to train question--answering systems~\cite{duan-etal-2017-question} or chatbots~\cite{lee-etal-2021-restatement}.

\section{Related Work}

While Question Generation is not as popular as the related task of Question Answering, there has been a steady increase in the number of publications in this area in recent years~\cite{amidei-etal-2018-evaluation,article}. Traditionally, rules and templates have been used to generate questions~\cite{mitkov-ha-2003-computer}; however, with the rise in popularity of deep neural networks, there was a shift towards using recurrent encored--decoder architectures~\cite{bao2020unilmv2,dong2019unified,du-etal-2017-learning,qi2020prophetnet,song-etal-2018-leveraging,xiao2020erniegen,zhou2018neural} and large-scale Transformers~\cite{devlin2019bert,lan2020albert,lewis2019bart,liu2019roberta,raffel2020exploring}.

The task is often formulated as one of generating a question given a target answer and a document as an input. Datasets such as SQuAD1.1~\cite{rajpurkar2016squad} and NewsQA~\cite{trischler-etal-2017-newsqa} are most commonly used for training, and the results are typically evaluated using measures such as BLEU~\cite{papineli2002bleu}, ROUGE~\cite{lin2004rouge}, and METEOR~\cite{lavie-agarwal-2007-meteor}. Note that this task formulation requires the target answer to be provided beforehand, which may not be practical for real-world situations. To get over this limitation, some systems extract all nouns and named entities from the input text as target answers, while other systems train a classifier to label all word $n$-grams from the text and to pick the ones with the highest probability to be answers~\cite{vachev2021generating}. To create context-related wrong options (i.e.,~distractors), typically the RACE dataset~\cite{lai-etal-2017-race} has been used along with beam search~\cite{chung-etal-2020-bert,gao2019generating,offerijns2020better}. 
Note that MOOCs pose additional challenges as they often cover specialized content that goes beyond knowledge found in Wikipedia, and can be offered in many languages; there are some open datasets that offer such kinds of questions in English~\cite{clark2018think,clark2019f,lai-etal-2017-race,mihaylov-etal-2018-suit,tafjord2019quarel} and in other languages~\cite{clark-etal-2020-tydi,hardalov-etal-2019-beyond,hardalov-etal-2020-exams,pmlr-v119-hu20b,jing-etal-2019-bipar,lewis-etal-2020-mlqa,lin2021few,9247161}.

Various practical systems have been developed for question generation. WebExperimenter~\cite{hoshino-nakagawa-2005-webexperimenter} generates Cloze-style questions for English proficiency testing. AnswerQuest~\cite{roemmele-etal-2021-answerquest} generates questions for better use in Question Answering systems, and  SQUASH~\cite{krishna2019generating} decomposes larger articles into paragraphs and generates a text comprehension question for each one; however, both systems lack the ability to generate distractors. There are also online services tailored to teachers. For example, Quillionz~\cite{quillionz} takes longer educational texts and generates questions according to a user-selected domain, while Questgen~\cite{questgen} can work with texts up to 500 words long. While these systems offer useful question recommendations, they also require paid licenses.
Our Leaf system offers a similar functionality, but is free and open-source, and can generate high-quality distractors. It is trained on publicly available data, and we are releasing our training scripts, thus allowing anybody to adapt the system to their own data.

\section{System}

\begin{figure}
\centering
\includegraphics[width=0.95\textwidth]{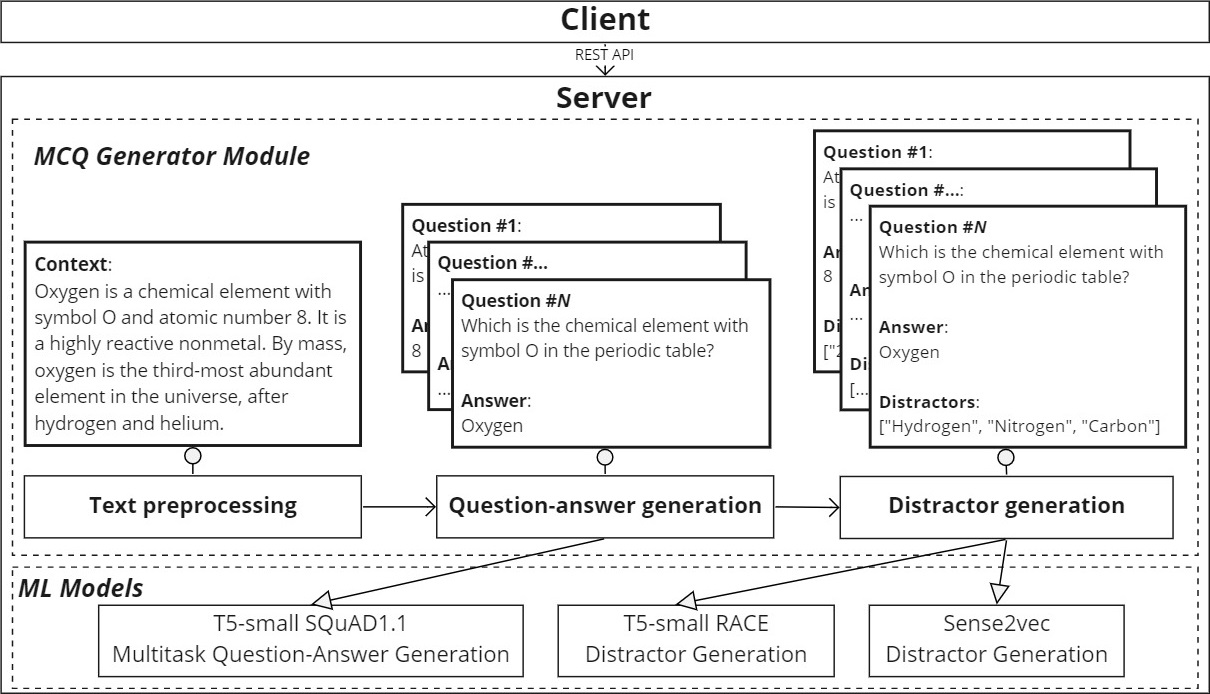}
\caption{The general architecture of Leaf.}
\label{fig:architecture}
\end{figure}

\textbf{System architecture:} Leaf has three main modules as shown in Figure~\ref{fig:architecture}. Using the \textit{Client}, an instructor inputs a required number of questions and her educational text. The text is then passed through a REST API to the \textit{Multiple-Choice Question (MCQ) Generator Module}, which performs pre-processing and then generates and returns the required number of question--answer pairs with distractors. To achieve higher flexibility and abstraction, the models implement an interface that allows them to be easily replaced.

\textbf{Question and Answer Generation:} To create the question--answer pairs, we combined the two tasks into a single multi-task model. We fine-tuned the small version of the T5 Transformer, which has 220M parameters, and we used the SQuAD1.1 dataset~\cite{rajpurkar2016squad}, which includes 100,000 question--answer pairs. We trained the model to output the question and the answer and to accept the passage and the answer with a 30\% probability for the answer to be replaced by the \texttt{[MASK]} token. This allows us to generate an answer for the input question by providing the \texttt{[MASK]} token instead of the target answer.
We trained the model for five epochs, and we achieved the best validation cross-entropy loss of 1.17 in the fourth epoch. We used a learning rate of 0.0001, a batch size of 16, and a source and a target maximum token lengths of 300 and 80, respectively. For question generation, we used the same data split and evaluation scripts as in~\cite{du-etal-2017-learning}. 
For answer generation, we trained on the modified SQuAD1.1 Question Answering dataset as proposed in our previous work~\cite{vachev2021generating}, achieving an Exact Match of 41.51 and an F1 score of 53.26 on the development set.

\textbf{Distractor Generation:} To create contextual distractors for the question--answer pairs, we used the RACE dataset~\cite{lai-etal-2017-race} and the small pre-trained T5 model. We provided the question, the answer, and the context as an input, and obtained three distractors separated by a \texttt{[SEP]} token as an output. We trained the model for five epochs, achieving a validation cross-entropy loss of 2.19. We used a learning rate of 0.0001, a batch size of 16, and a source and a target maximum token lengths of 512 and 64, respectively. The first, the second, and the third distractor had BLEU1 scores of 46.37, 32.19, and 34.47, respectively. 
We further extended the variety of distractors with context-independent proposals, using  sense2vec~\cite{trask2015sense2vec} to generate words or multi-word phrases that are semantically similar to the answer.

\begin{figure}
\centering
\includegraphics[width=\textwidth]{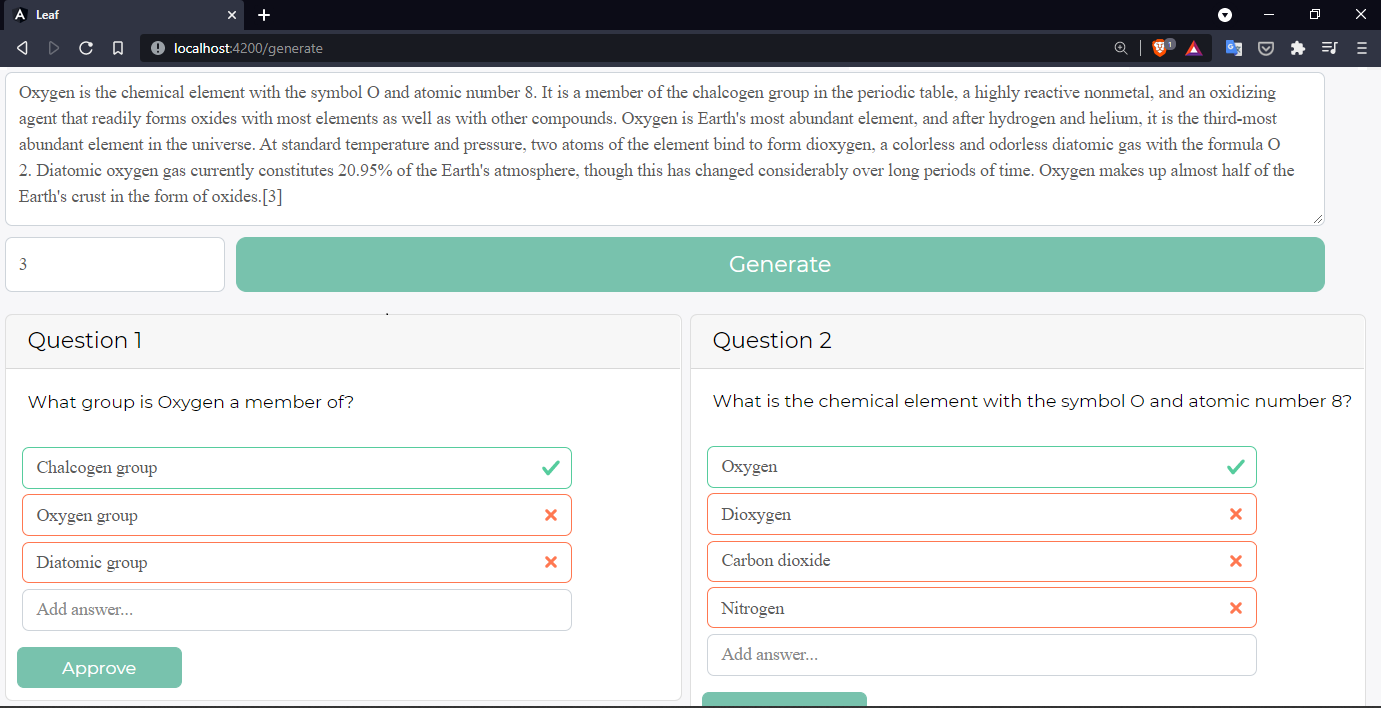}
\caption{Screenshot of Leaf showing the generated questions for a passage from the Wikipedia article on Oxygen. All distractors in \textit{Question 1} are generated by the T5 model, and the last two distractors in \textit{Question 2} are generated by the sense2vec model.}

\label{fig:ui}
\end{figure}

\textbf{User Interface:} Using the user interface shown on Figure~\ref{fig:ui}, the instructor can input her educational text, together with the desired number of questions to generate. Then, she can choose some of them, and potentially edit them, before using them as part of her course.

\section{Conclusion and Future Work}

We presented Leaf, a system to generate multiple-choice questions from text. The system can be used both in the classroom and in an industrial setting to detect knowledge gaps or as a self-assessment tool; it could also be integrated as part of other systems. With the aim to enable a better educational process, especially in the context of MOOCs, we open-source the project, including all training scripts and documentation.

In future work, we plan to experiment with a variety of larger pre-trained Transformers as the underlying model. We further plan to train on additional data. Given the lack of datasets created specifically for the task of Question Generation, we plan to produce a new dataset by using Leaf in real university courses and then collecting and manually curating the question--answer pairs Leaf generates over time.

\section*{Acknowledgements}
This research is partially supported by Project UNITe BG05M2OP001-1.001-0004 funded by the Bulgarian OP ``Science and Education for Smart Growth.''

\bibliographystyle{plain}
\bibliography{bibliography}

\end{document}